\documentclass{article}

\usepackage{arxiv}

\usepackage[utf8]{inputenc} 
\usepackage[T1]{fontenc}    
\usepackage{hyperref}       
\usepackage{url}            
\usepackage{booktabs}       
\usepackage{amsfonts}       
\usepackage{nicefrac}       
\usepackage{microtype}      
\usepackage[pdftex]{graphicx}
\usepackage{amssymb,amsfonts,amsmath,amscd}
\usepackage[printonlyused,withpage]{acronym}
\usepackage{natbib}
\usepackage{mathrsfs}
\usepackage{rotating}

\graphicspath{{figs/}}

\newcommand{\bbm}{\begin{bmatrix}}
\newcommand{\ebm}{\end{bmatrix}}
\newcommand{\mbf}{\mathbf}
\newcommand{\mbs}[1]{{\boldsymbol{#1}}}

\newcommand{\beq}{\begin{equation}}
\newcommand{\eeq}{\end{equation}}
\newcommand{\bdis}{\begin{displaymath}}
\newcommand{\edis}{\end{displaymath}}
\newcommand{\beqn}[1]{\begin{subequations}\label{eq:#1}\begin{eqnarray}}
\newcommand{\eeqn}{\end{eqnarray}\end{subequations}}
\acrodef{BA}{Bundle Adjustment}
\acrodef{DNN}{Deep Neural Network}
\acrodef{EM}{Expectation Maximization}
\acrodef{GEM}{Generalized Expectation Maximization}
\acrodef{HMM}{Hidden Markov Model}
\acrodef{LDS}{Linear Dynamical System}
\acrodef{LQG}{Linear Quadratic Gaussian}
\acrodef{LQR}{Linear Quadratic Regulator}
\acrodef{LTI}{Linear Time-Invariant}
\acrodef{RTS}{Rauch-Tung-Striebel}
\acrodef{SGD}{Stochastic Gradient Descent}
\acrodef{SLAM}{Simultaneous Localization and Mapping}
\acrodef{GVI}{Gaussian Variational Inference}
\acrodef{ESGVI}{Exactly Sparse Gaussian Variational Inference}
\acrodef{MAP}{Maximum A Posteriori}
\acrodef{ML}{Maximum Likelihood}
\acrodef{KL}{Kullback-Leibler}
\acrodef{PDF}{Probability Density Function}
\acrodef{NEES}{Normalized Estimation Squared Error}
\acrodef{KF}{Kalman Filter}
\acrodef{ISPKF}{Iterated Sigmapoint Kalman Filter}
\acrodef{ESGVI-GN}{ESGVI Gauss-Newton}
\acrodef{ELBO}{Expectation Lower Bound}
\acrodef{NGD}{Natural Gradient Descent}
\acrodef{FIM}{Fisher Information Matrix}
\acrodef{RANSAC}{Random Sample And Consensus}

\hypersetup{%
    pdftitle={},
    pdfauthor={},
    pdfkeywords={},
    pdfsubject={},
    pdfstartview=FitH,%
    bookmarks=true,%
    bookmarksopen=true,%
    breaklinks=true,%
    colorlinks=true,%
    linkcolor=black,
    anchorcolor=black,%
    citecolor=black,
    filecolor=black,%
    menucolor=black,
    pagecolor=black,%
    urlcolor=black
}%

\title{Multivariate Gaussian Variational Inference \\ by Natural Gradient Descent}

\author{
 {\normalfont Timothy D. Barfoot} \\
 Institute for Aerospace Studies \\
 University of Toronto \\
 \texttt{tim.barfoot@utoronto.ca} 
}

\begin{document}

\maketitle
\title{Multivariate Gaussian Variational Inference by Natural Gradient Descent}

\begin{abstract}
This short note reviews so-called \ac{NGD} for multivariate Gaussians.  The \ac{FIM} is derived for several different parameterizations of Gaussians.  Careful attention is paid to the symmetric nature of the covariance matrix when calculating derivatives.  We show that there are some advantages to choosing a parameterization comprising the mean and inverse covariance matrix and provide a simple \ac{NGD} update that accounts for the symmetric (and sparse) nature of the inverse covariance matrix.
\end{abstract}

\keywords{Gaussian variational inference \and natural gradient descent \and Kronecker algebra \and Fisher information matrix}

\section{Mathematical Tools}

\renewcommand{\vec}{\mbox{vec}}
\newcommand{\vech}{\mbox{vech}}
\newcommand{\mat}{\mbox{mat}}
\newcommand{\matf}{\mbox{matf}}
\newcommand{\sym}{\mbox{sym}}
\newcommand{\tr}{\mbox{tr}}
\newcommand{\rank}{\mbox{rank}}

This first section introduces some less common tools that make it possible to take derivatives of general expressions of symmetric matrices.  Readers familiar with these already can jump to the next section.

\subsection{Vectorization}

There are several identities involving the Kronecker product, $\otimes$, and $\vec(\cdot)$ operator (that stacks the columns of a matrix) of which we will make use:
\begin{subequations}
\begin{eqnarray}
\vec(\mbf{a}) & \equiv & \mbf{a} \\
\vec(\mbf{a}\mbf{b}^T) & \equiv & \mbf{b} \otimes \mbf{a} \\
\vec(\mbf{A}\mbf{B}\mbf{C}) & \equiv & (\mbf{C}^T \otimes \mbf{A} )\, \vec(\mbf{B}) \\
\vec(\mbf{A})^T \vec(\mbf{B}) & \equiv & \tr(\mbf{A}^T\mbf{B}) \\
(\mbf{A} \otimes \mbf{B})(\mbf{C} \otimes \mbf{D}) & \equiv & (\mbf{A}\mbf{C}) \otimes (\mbf{B}\mbf{D}) \\
(\mbf{A} \otimes \mbf{B})^{-1} & \equiv & \mbf{A}^{-1} \otimes \mbf{B}^{-1} \\
 (\mbf{A} \otimes \mbf{B})^{T} & \equiv & \mbf{A}^{T} \otimes \mbf{B}^{T} \\
 \left| \mbf{A}_{N \times N} \otimes \mbf{B}_{M \times M} \right| & \equiv & |\mbf{A}|^M |\mbf{B}|^N \\ 
 \rank(\mbf{A} \otimes \mbf{B}) & \equiv & \rank(\mbf{A}) \, \rank(\mbf{B}) \\
 \tr(\mbf{A} \otimes \mbf{B}) & \equiv & \tr(\mbf{A}) \, \tr(\mbf{B}) \\
 \mbf{a}^T \mbf{B} \mbf{C} \mbf{B}^T \mbf{d} & \equiv & \vec(\mbf{B})^T ( \mbf{C} \otimes \mbf{d}\mbf{a}^T ) \vec(\mbf{B}) 
\end{eqnarray}
\end{subequations}
We shall also define the $\mat(\cdot)$ operator to mean the inverse of the $\vec(\cdot)$ operator; in other words it unstacks the columns back into the original matrix, whose original size is presumably remembered somehow.  Thus
\begin{equation}
\mat\left( \vec\left(\mbf{A} \right) \right) \equiv \mbf{A}.
\end{equation}
It is worth noting that $\otimes$, $\vec(\cdot)$, and $\mat(\cdot)$ are all linear operators.  

\subsection{Matrix Calculus Using Differentials}

To handle taking derivatives of complicated expressions involving matrices, we follow the approach of \citet[\S 18]{magnus19}.  This section discusses (unconstrained) differentials before moving on to how to handle symmetric matrices.  Some of the less usual results that we will make use of include
\begin{subequations}
\begin{eqnarray}
d\tr(\mbf{X}) & = & \tr( d\mbf{X} ) \\
d | \mbf{X}| & = & | \mbf{X}| \, \tr\left( \mbf{X}^{-1} \,d\mbf{X} \right) \\
d \ln | \mbf{X} | & = & \tr\left( \mbf{X}^{-1} \,d\mbf{X} \right) \\
d \mbf{X}^{-1} & = & - \mbf{X}^{-1} \, d\mbf{X} \, \mbf{X}^{-1} \\
d  f( \mbf{X} ) & = & \tr\left(\frac{\partial f}{\partial \mbf{X}} \, d\mbf{X} \right)
\end{eqnarray}
\end{subequations}
All of the usual linear operations for differentials apply as well.  From the last relationship, we see that if we can manipulate our differential into the form
\begin{equation}
d  f( \mbf{X} ) = \tr\left(\mbf{A} \, d\mbf{X} \right),
\end{equation}
we can read the Jacobian matrix,
\begin{equation}
\mbf{A} = \frac{\partial f}{\partial \mbf{X}},
\end{equation}
directly.  Another way to see this is to make use of vectorization.  We can rewrite the differential as
\begin{equation}
d  f( \mbf{X} ) = \vec\left(\mbf{A}\right)^T \vec\left(d\mbf{X} \right) = d\vec\left(\mbf{X} \right)^T \vec\left(\mbf{A}\right), 
\end{equation}
so that
\begin{equation}
\frac{\partial f(\mbf{X})}{\partial \vec\left(\mbf{X} \right)^T} = \vec\left(\mbf{A}\right) .
\end{equation}
Then, converting back to a matrix we have
\begin{equation}
\frac{\partial f}{\partial \mbf{X}} = \mat\left( \frac{\partial f(\mbf{X})}{\partial \vec\left(\mbf{X} \right)^T} \right) = \mat\left( \vec\left(\mbf{A}\right) \right) =\mbf{A}.
\end{equation}
These expressions can be used recursively to calculate second differentials as well.  The main idea to minimize tedious calculations is to first build all the differentials then use the vectorization tools from the previous section to assemble them into a Jacobian and/or Hessian. 

\subsection{Parameterizing Symmetric Matrices Without Duplication}

In the previous section, we introduced tools to calculate the first and second derivatives of expressions with respect to a matrix.  However, if the matrix is symmetric, we need to modify the results because the elements above and below the main diagonal are duplicated.  We again follow the approach of \citet[\S 18]{magnus19}.

We begin by introducing the $\vech(\cdot)$, operator\footnote{Presumably the addition of `h' indicates the lower `half'.} that stacks up the elements in a matrix, excluding all the elements above the main diagonal.  Then, we define the {\em duplication matrix}, $\mbf{D}$, that allows us to build the full symmetric matrix from its unique parts:
\begin{equation}
\vec(\mbf{A}) = \mbf{D} \, \vech(\mbf{A})  \qquad \mbox{(symmetric $\mbf{A}$)}
\end{equation}
It is useful to consider a simple $2 \times 2$ example:
\begin{equation}
\mbf{A} = \bbm a & b \\ b & c \ebm, \quad \vec(\mbf{A}) = \bbm a \\ b \\ b \\ c \ebm, \quad \mbf{D} = \bbm 1 & 0 & 0 \\ 0 & 1 & 0 \\ 0 & 1 & 0 \\ 0 & 0 & 1 \ebm , \quad \vech(\mbf{A}) = \bbm a \\ b \\ c \ebm.
\end{equation}
If we want to convert back to a matrix it is useful to define a corresponding $\matf(\cdot)$ operator\footnote{The `f' indicates we are converting a half vector back into a `full' symmetric matrix.} so that
\begin{equation}
\matf(\vech(\mbf{A})) = \mat(\mbf{D} \, \vech(\mbf{A})) = \mat( \vec(\mbf{A}) ) = \mbf{A} \qquad \mbox{(symmetric $\mbf{A}$)}.
\end{equation}
The {\em Moore-Penrose pseudoinverse} of $\mbf{D}$ will be denoted $\mbf{D}^+$ and is given by
\begin{equation}
\mbf{D}^+ = \left( \mbf{D}^T \mbf{D} \right)^{-1} \mbf{D}^T.
\end{equation}
We can then use $\mbf{D}^+$ to calculate the unique vector from the nonunique vector:
\begin{equation}
\vech(\mbf{A}) = \mbf{D}^+ \vec(\mbf{A})  \qquad \mbox{(symmetric $\mbf{A}$)}.
\end{equation}
For our $2 \times 2$ example we have
\begin{equation}
\mbf{D}^+ = \bbm 1 & 0 & 0 & 0 \\ 0 & \frac{1}{2} & \frac{1}{2} & 0 \\ 0 & 0 & 0 & 1 \ebm.
\end{equation}
Useful identities involving $\mbf{D}$ are then
\begin{subequations}
\label{eq:dup}
\begin{eqnarray}
\mbf{D}^+ \mbf{D} & \equiv & \mbf{1} \label{eq:dup1} \\
\mbf{D}^{+^T} \mbf{D}^T & \equiv & \mbf{D} \mbf{D}^+ \label{eq:dup2} \\
\mbf{D} \mbf{D}^+ \vec(\mbf{A}) & \equiv & \vec(\mbf{A})  \qquad\qquad\!\mbox{(symmetric $\mbf{A}$)} \label{eq:dup3} \\
\mbf{D} \mbf{D}^+ \left( \mbf{A} \otimes \mbf{A} \right) \mbf{D} & \equiv & \left( \mbf{A} \otimes \mbf{A} \right) \mbf{D}  \qquad \mbox{(any $\mbf{A}$)} \label{eq:dup4}
\end{eqnarray}
\end{subequations}
which can be found in \citet{magnus80}.  We can also define the $\sym(\cdot)$ operator as
\begin{equation}
\sym(\mbf{A}) = \mbf{A} + \mbf{A}^T - \mbf{A} \circ \mbf{1},
\end{equation}
with $\circ$ the Hadamard (elementwise) product.  Then we can relate the $\sym(\cdot)$ operator to the duplication matrix as follows:
\begin{equation}
\mbf{D} \mbf{D}^T \vec(\mbf{A}) = \vec\left( \sym( \mbf{A} ) \right),
\end{equation}
which holds for any $\mbf{A}$.

\subsection{Differentials for Symmetric Matrices}

It is now straightforward to calculate derivatives of functions of symmetric matrices.  The key idea is to use the symmetry-aware parameterization from the last section:
\begin{equation}
\vec(\mbf{A}) = \mbf{D} \, \vech(\mbf{A}).
\end{equation}
Taking the differential of this we have
\begin{equation}
d\vec(\mbf{A}) = \mbf{D} \, d\vech(\mbf{A}).
\end{equation}
We can insert this whenever we have a differential involving a symmetric matrix:
\begin{equation}
d  f( \mbf{X} ) = d\vec\left(\mbf{X} \right)^T \vec\left(\frac{\partial f}{\partial \mbf{X}}\right) = d\vech\left(\mbf{X} \right)^T \mbf{D}^T \vec\left(\frac{\partial f}{\partial \mbf{X}}\right),
\end{equation}
so that
\begin{equation}
\frac{\partial f(\mbf{X})}{\partial \vech\left(\mbf{X} \right))^T} = \mbf{D}^T \vec\left(\frac{\partial f}{\partial \mbf{X}}\right).
\end{equation}
What the extra $\mbf{D}^T$ effectively does is add together the unconstrained elements of the derivative corresponding to the same element above and below the diagonal and then maps this to a single parameter in the unique representation.  

If we define $\eth$ to indicate a partial derivative with respect to a full symmetric matrix (where we have accounted for the symmetry) we can write
\begin{multline}
\frac{\eth f}{\eth \mbf{X}} = \matf\left( \frac{\partial f(\mbf{X})}{\partial \vech\left(\mbf{X} \right))^T} \right) = \mat\left( \mbf{D} \frac{\partial f(\mbf{X})}{\partial \vech\left(\mbf{X} \right))^T}\right) = \mat\left( \mbf{D} \mbf{D}^T \vec\left(\frac{\partial f}{\partial \mbf{X}}\right) \right) \\ = \mat\left( \vec\left( \sym\left( \frac{\partial f}{\partial \mbf{X}} \right) \right) \right) = \sym\left( \frac{\partial f}{\partial \mbf{X}}\right),
\end{multline}
which is now in terms of the $\sym(\cdot)$ operator.


\newpage
\section{Fisher Information Matrix for a Multivariate Gaussian}

To carry out \ac{NGD}, we will have need of the \acf{FIM} for a multivariate Gaussian.  However, there are several useful ways to parameterize a Gaussian, each with its own \ac{FIM}, so we will show a few.  First we will derive the general expression.

\subsection{\ac{FIM} Derivation}

A multivariate Gaussian \ac{PDF} takes the form
\begin{equation}
q(\mbf{x}) = \mathcal{N}(\mbs{\mu}, \mbs{\Sigma}) = \frac{1}{\sqrt{(2\pi)^N |\mbs{\Sigma}|}} \exp\left( -\frac{1}{2} (\mbf{x} - \mbs{\mu})^T \mbs{\Sigma}^{-1} (\mbf{x} - \mbs{\mu}) \right),
\end{equation}
which has been parameterized using mean, $\mbs{\mu}$, and covariance, $\mbs{\Sigma}$.   

We will use the \ac{KL} divergence \citep{kullback51} to define the \acf{FIM} \citep{fisher22}.  The KL divergence between two Gaussians, $q$ and $q^\prime$, can be expressed as
\begin{equation}
\mbox{KL}(q||q^\prime) = - \int q(\mbf{x}) \ln \left( \frac{q^\prime(\mbf{x})}{q(\mbf{x})} \right) d\mbf{x} = \mathbb{E}_q \left[ \ln q(\mbf{x}) - \ln q^\prime(\mbf{x}) \right].
\end{equation}
If we suppose that $q$ and $q^\prime$ are infinitesimally close to one another in some parameter space then
\begin{equation}
\ln q^\prime(\mbf{x}) \approx \ln q(\mbf{x}) + d\ln q(\mbf{x}) + \frac{1}{2} d^2\ln q(\mbf{x}) ,
\end{equation}
and so
\begin{equation}
\label{eq:kl2gaussians}
\mbox{KL}(q||q^\prime) \approx \mathbb{E}_q \left[ -d\ln q(\mbf{x}) - \frac{1}{2} d^2\ln q(\mbf{x}) \right],
\end{equation}
out to second order in the differentials.  For Gaussians (and some other distributions) the first term is in fact zero.  To see this, we write the negative log-likelihood of $q(\mbf{x})$ as
\begin{equation}
-\ln q(\mbf{x}) = \frac{1}{2} (\mbf{x} - \mbs{\mu})^T \mbs{\Sigma}^{-1} (\mbf{x} - \mbs{\mu}) + \frac{1}{2} \ln | \mbs{\Sigma} | + \mbox{constant}.
\end{equation}
The first differential is
\begin{multline}
- d \ln q(\mbf{x}) = -d\mbs{\mu}^T \mbs{\Sigma}^{-1} (\mbf{x} - \mbs{\mu}) + \frac{1}{2}  (\mbf{x} - \mbs{\mu})^T d\mbs{\Sigma}^{-1} (\mbf{x} - \mbs{\mu})  + \frac{1}{2} \tr\left( \mbs{\Sigma}^{-1} \, d\mbs{\Sigma} \right) \\ = -d\mbs{\mu}^T \mbs{\Sigma}^{-1} (\mbf{x} - \mbs{\mu})  + \frac{1}{2} \tr\left(  \left( \mbs{\Sigma}^{-1} - \mbs{\Sigma}^{-1}(\mbf{x} - \mbs{\mu})(\mbf{x} - \mbs{\mu})^T\mbs{\Sigma}^{-1} \right) \, d\mbs{\Sigma} \right),
\end{multline}
and so 
\begin{equation}
\mathbb{E}_q \left[ - d \ln q(\mbf{x}) \right] = -d\mbs{\mu}^T \mbs{\Sigma}^{-1} \underbrace{\mathbb{E}[\mbf{x} - \mbs{\mu}]}_{\mbf{0}}  + \frac{1}{2} \tr\biggl(  \bigl( ( \mbs{\Sigma}^{-1} - \mbs{\Sigma}^{-1} \underbrace{\mathbb{E}\left[(\mbf{x} - \mbs{\mu})(\mbf{x} - \mbs{\mu})^T\right]}_{\mbs{\Sigma}}\mbs{\Sigma}^{-1} \bigr)) \, d\mbs{\Sigma} \biggr) = \mbf{0}.
\end{equation}
Turning the differentials into partial derivatives, we can rewrite~\eqref{eq:kl2gaussians} as
\begin{equation}
\mbox{KL}(q||q^\prime) \approx \mathbb{E}_q \left[ - \frac{1}{2} d^2\ln q(\mbf{x}) \right] = \frac{1}{2} \delta\mbs{\theta}^T  \underbrace{\mathbb{E}_q \left[ -\frac{\partial^2\ln q(\mbf{x})}{\partial \mbs{\theta}^T \partial \mbs{\theta}} \right]}_{\mbs{\mathcal{I}}_{\mbs{\theta}}} \, \delta\mbs{\theta},
\end{equation}
for some parameterization, $\mbs{\theta}$, of a Gaussian.  The matrix, $\mbs{\mathcal{I}}_{\mbs{\theta}}$, is called the \acf{FIM} and defines a Riemmannian metric tensor for the parameters, $\mbs{\theta}$.  As a preview of what is to come, for cost functions that are similar to \ac{KL} divergence, we will use the \ac{FIM} as an approximation of the Hessian to build a Newton-like optimizer.  We will next work out the \ac{FIM} for a few parameterizations of Gaussians.

\subsection{Canonical Parameterization - Not Accounting for Symmetry}

We begin with the most obvious parameterization, the mean and (vectorized) covariance:
\begin{equation}
\mbs{\theta} = \bbm \mbs{\mu} \\ \vec(\mbs{\Sigma}) \ebm,
\end{equation}
which we will refer to as the {\em canonical parameterization}.  To calculate the \ac{FIM}, we need the second differential,
\begin{multline}
- d^2 \ln q(\mbf{x}) = d\mbs{\mu}^T \mbs{\Sigma}^{-1} \,d\mbs{\mu} - 2 d\mbs{\mu}^T \mbs{\Sigma}^{-1} (\mbf{x} - \mbs{\mu})  + \frac{1}{2} \tr\left(  \mbs{\Sigma}^{-1} \, d^2\mbs{\Sigma} \, \mbs{\Sigma}^{-1} \left( \mbs{\Sigma} - (\mbf{x}-\mbs{\mu}) (\mbf{x}-\mbs{\mu})^T \right) \right) \\  \frac{1}{2} \tr\left(  \mbs{\Sigma}^{-1} \, d\mbs{\Sigma} \, \mbs{\Sigma}^{-1} \, d\mbs{\Sigma} \, \mbs{\Sigma}^{-1} \left(  (\mbf{x}-\mbs{\mu}) (\mbf{x}-\mbs{\mu})^T - \frac{1}{2} \mbs{\Sigma} \right) \right).
\end{multline}
The expected value of the second differential over $q(\mbf{x})$ is
\begin{equation}
-\mathbb{E}_q \left[ d^2 \ln q(\mbf{x}) \right] = d\mbs{\mu}^T \mbs{\Sigma}^{-1} \,d\mbs{\mu} + \frac{1}{2} \tr\left(  \mbs{\Sigma}^{-1} \, d\mbs{\Sigma} \, \mbs{\Sigma}^{-1} \, d\mbs{\Sigma} \right).
\end{equation}
Vectorizing $\mbs{\Sigma}$ we have the nice result
\begin{equation}
\label{eq:seconddiff}
-\mathbb{E}_q \left[ d^2 \ln q(\mbf{x}) \right] = d\mbs{\theta}^T \underbrace{\bbm \mbs{\Sigma}^{-1} & \mbf{0} \\ \mbf{0} & \frac{1}{2} \left(\mbs{\Sigma}^{-1} \otimes \mbs{\Sigma}^{-1}\right) \ebm}_{\mbs{\mathcal{I}}_{\mbs{\theta}}} d\mbs{\theta}.
\end{equation}
The inverse \ac{FIM} is simply
\begin{equation}
\mbs{\mathcal{I}}_{\mbs{\theta}}^{-1} = \bbm \mbs{\Sigma} & \mbf{0} \\ \mbf{0} & 2 \left(\mbs{\Sigma} \otimes \mbs{\Sigma} \right) \ebm.
\end{equation}
The trouble is that we have not accounted for the symmetric nature of $\mbs{\Sigma}$.  

\subsection{Canonical Parameterization - Accounting for Symmetry}

To account for the symmetric nature of $\mbs{\Sigma}$, we define the symmetry-aware parameterization as
\begin{equation}
\mbs{\gamma} = \bbm \mbs{\gamma}_1 \\ \mbs{\gamma}_2 \ebm = \bbm \mbs{\mu} \\ \vech(\mbs{\Sigma}) \ebm.
\end{equation}
Then we have
\begin{equation}
d\mbs{\mu} = d\mbs{\gamma}_1, \quad \vec\left(d\mbs{\Sigma}\right) = \mbf{D} \, d\mbs{\gamma}_2,
\end{equation}
or
\begin{equation}
d\mbs{\theta} = \bbm \mbf{1} & \mbf{0} \\ \mbf{0} & \mbf{D} \ebm  d\mbs{\gamma}.
\end{equation}
Substituting this into~\eqref{eq:seconddiff}, we have
\begin{equation}
-\mathbb{E}_q \left[ d^2 \ln q(\mbf{x}) \right] = d\mbs{\gamma}^T \underbrace{\bbm \mbs{\Sigma}^{-1} & \mbf{0} \\ \mbf{0} & \frac{1}{2} \mbf{D}^T \left(\mbs{\Sigma}^{-1} \otimes \mbs{\Sigma}^{-1}\right)\mbf{D} \ebm}_{\mbs{\mathcal{I}}_{\mbs{\gamma}}}  d\mbs{\gamma}.
\end{equation}
The inverse \ac{FIM} is given by
\begin{equation}
\mbs{\mathcal{I}}_{\mbs{\gamma}}^{-1} = \bbm \mbs{\Sigma} & \mbf{0} \\ \mbf{0} & 2 \mbf{D}^+ \left(\mbs{\Sigma} \otimes \mbs{\Sigma}\right)\mbf{D}^{+^T} \ebm,
\end{equation}
corresponding to \citet[\S 18]{magnus19}.

\subsection{Hybrid Parameterization - Not Accounting for Symmetry}

In many real large-scale problems, the inverse covariance matrix is sparse so that it is much more desirable to work with it directly.  We call this the {\em hybrid} parameterization as the mean is as before while we use the inverse covariance:
\begin{equation}
\mbs{\alpha} = \bbm \mbs{\alpha}_1 \\ \mbs{\alpha}_2 \ebm = \bbm \mbs{\mu} \\ \vec(\mbs{\Sigma}^{-1}) \ebm.
\end{equation} 
We then have
\begin{equation}
d\mbs{\mu} = d\mbs{\alpha}_1, \quad \vec\left( d\mbs{\Sigma}^{-1}\right) =  d\mbs{\alpha}_2.
\end{equation}
Expanding the second of these we see
\begin{equation}
d\mbs{\alpha}_2 = \vec\left(d\mbs{\Sigma}^{-1}\right) = \vec\left(-\mbs{\Sigma}^{-1}\, d\mbs{\Sigma} \, \mbs{\Sigma}^{-1} \right) = - \left( \mbs{\Sigma}^{-1} \otimes \mbs{\Sigma}^{-1} \right) \vec\left( d\mbs{\Sigma} \right),
\end{equation}
and so
\begin{equation}
\vec\left( d\mbs{\Sigma} \right) = - \left( \mbs{\Sigma} \otimes \mbs{\Sigma} \right) d\mbs{\alpha}_2.
\end{equation}
We therefore have
\begin{equation}
d\mbs{\theta} = \bbm \mbf{1} & \mbf{0} \\ \mbf{0} & - \left( \mbs{\Sigma} \otimes \mbs{\Sigma} \right)  \ebm  d\mbs{\alpha}.
\end{equation}
Substituting this into~\eqref{eq:seconddiff} we have
\begin{equation}
-\mathbb{E}_q \left[ d^2 \ln q(\mbf{x}) \right] = d\mbs{\alpha}^T \underbrace{\bbm \mbs{\Sigma}^{-1} & \mbf{0} \\ \mbf{0} & \frac{1}{2}  \left(\mbs{\Sigma} \otimes \mbs{\Sigma}\right) \ebm}_{\mbs{\mathcal{I}}_{\mbs{\alpha}}} d\mbs{\alpha}.
\end{equation}
The inverse \ac{FIM} is given by
\begin{equation}
\mbs{\mathcal{I}}_{\mbs{\alpha}}^{-1} = \bbm \mbs{\Sigma} & \mbf{0} \\ \mbf{0} & 2 \left(\mbs{\Sigma}^{-1} \otimes \mbs{\Sigma}^{-1}\right) \ebm,
\end{equation}
which follows a similar form to the previous section.

\subsection{Hybrid Parameterization - Accounting for Symmetry}

If we want to account for symmetry we choose
\begin{equation}
\mbs{\beta} = \bbm \mbs{\beta}_1 \\ \mbs{\beta}_2 \ebm = \bbm \mbs{\mu} \\ \vech(\mbs{\Sigma}^{-1}) \ebm.
\end{equation} 
Similarly to how we convert to the symmetry-aware version of the canonical representation, we have
\begin{equation}
d\mbs{\beta} = \bbm \mbf{1} & \mbf{0} \\ \mbf{0} & \mbf{D} \ebm  d\mbs{\alpha}.
\end{equation}
Substituting this into~\eqref{eq:seconddiff} we have
\begin{equation}
-\mathbb{E}_q \left[ d^2 \ln q(\mbf{x}) \right] = d\mbs{\beta}^T \underbrace{\bbm \mbs{\Sigma}^{-1} & \mbf{0} \\ \mbf{0} & \frac{1}{2} \mbf{D}^T \left(\mbs{\Sigma} \otimes \mbs{\Sigma}\right)\mbf{D} \ebm}_{\mbs{\mathcal{I}}_{\mbs{\beta}}} d\mbs{\beta}.
\end{equation}
The inverse \ac{FIM} is given by
\begin{equation}
\mbs{\mathcal{I}}_{\mbs{\beta}}^{-1} = \bbm \mbs{\Sigma} & \mbf{0} \\ \mbf{0} & 2 \mbf{D}^+ \left(\mbs{\Sigma}^{-1} \otimes \mbs{\Sigma}^{-1}\right)\mbf{D}^{+^T} \ebm,
\end{equation}
which follows a similar form to the previous section.

\subsection{Natural (or Inverse Covariance Form) Parameterization - Not Accounting for Symmetry}

Somewhat confusingly named\footnote{There does not seem to be a connection with {\em natural} gradient descent.}, the {\em natural parameters} for a Gaussian can be defined as
\begin{equation}
\mbs{\eta} = \bbm \mbs{\Sigma}^{-1} \mbs{\mu} \\ \vec\left(\mbs{\Sigma}^{-1}\right) \ebm.
\end{equation}
This is also sometimes called the {\em inverse covariance form} of a Gaussian.  The derivation for these parameters is somewhat more involved as the \ac{FIM} is no longer block diagonal, making this choice somewhat {\em unnatural}.

We will build off the hybrid parameterization as it already deals with the inverse covariance matrix and so the covariance parameter differentials are the same, $d\mbs{\eta}_2 = d\mbs{\alpha}_2$.  The differential for the mean is
\begin{equation}
d\mbs{\eta}_1 = d\mbs{\Sigma}^{-1} \, \mbs{\mu} + \mbs{\Sigma}^{-1} \, d\mbs{\mu} =  \mbs{\Sigma}^{-1} \, d\mbs{\alpha}_1 + \left( \mbs{\mu}^T \otimes \mbf{1} \right) \, d\mbs{\alpha}_2,
\end{equation}
which takes a bit of manipulation.  Stacking these we have
\begin{equation}
d\mbs{\eta} = \bbm  \mbs{\Sigma}^{-1} &  \left( \mbs{\mu}^T \otimes \mbf{1} \right)  \\ \mbf{0} & \mbf{1} \ebm d\mbs{\alpha},
\end{equation}
which is the reverse relationship from what we want, but we will need this coefficient matrix when computing the inverse \ac{FIM}.  Inverting we have
\begin{equation}
d\mbs{\alpha} = \bbm  \mbs{\Sigma} &  -\mbs{\Sigma} \left( \mbs{\mu}^T \otimes \mbf{1} \right)  \\ \mbf{0} & \mbf{1} \ebm d\mbs{\eta}.
\end{equation}
The \ac{FIM} is then
\begin{equation}
\mbs{\mathcal{I}}_{\mbs{\eta}} =  \bbm  \mbs{\Sigma} &  -\mbs{\Sigma} \left( \mbs{\mu}^T \otimes \mbf{1} \right)  \\ \mbf{0} & \mbf{1} \ebm^T  \mbs{\mathcal{I}}_{\mbs{\alpha}} \bbm  \mbs{\Sigma} &  -\mbs{\Sigma} \left( \mbs{\mu}^T \otimes \mbf{1} \right)  \\ \mbf{0} & \mbf{1} \ebm = \bbm \mbs{\Sigma} & - \mbs{\Sigma} \left( \mbs{\mu}^T \otimes \mbf{1} \right) \\ - \left( \mbs{\mu} \otimes \mbf{1} \right)\mbs{\Sigma} & \frac{1}{2} \left( \mbs{\Sigma} \otimes \mbs{\Sigma} \right) + \left( \mbs{\mu} \otimes \mbf{1} \right) \mbs{\Sigma} \left( \mbs{\mu}^T \otimes \mbf{1} \right) \ebm.
\end{equation}
The inverse \ac{FIM} is then
\begin{multline}
\mbs{\mathcal{I}}_{\mbs{\eta}}^{-1} = \bbm  \mbs{\Sigma}^{-1} &  \left( \mbs{\mu}^T \otimes \mbf{1} \right)  \\ \mbf{0} & \mbf{1} \ebm  \mbs{\mathcal{I}}_{\mbs{\alpha}}^{-1}  \bbm  \mbs{\Sigma}^{-1} &  \left( \mbs{\mu}^T \otimes \mbf{1} \right)  \\ \mbf{0} & \mbf{1} \ebm^T  \\ = \bbm  \mbs{\Sigma}^{-1} + 2  \left( \mbs{\mu} \otimes \mbf{1} \right)  \left( \mbs{\Sigma}^{-1} \otimes \mbs{\Sigma}^{-1} \right)  \left( \mbs{\mu}^T \otimes \mbf{1} \right) & 2  \left( \mbs{\mu}^T \otimes \mbf{1} \right) \left( \mbs{\Sigma}^{-1} \otimes \mbs{\Sigma}^{-1} \right) \\ 2 \left( \mbs{\Sigma}^{-1} \otimes \mbs{\Sigma}^{-1} \right) \left( \mbs{\mu} \otimes \mbf{1} \right) & 2 \left( \mbs{\Sigma}^{-1} \otimes \mbs{\Sigma}^{-1} \right)\ebm \\ = \bbm \left( 1 + 2 \mbs{\mu}^T \mbs{\Sigma}^{-1} \mbs{\mu} \right) \mbs{\Sigma}^{-1} & 2 \left( \mbs{\mu}^T \mbs{\Sigma}^{-1} \otimes \mbs{\Sigma}^{-1} \right) \\ 2 \left(  \mbs{\Sigma}^{-1} \mbs{\mu} \otimes \mbs{\Sigma}^{-1} \right) & 2 \left( \mbs{\Sigma}^{-1} \otimes \mbs{\Sigma}^{-1} \right) \ebm, 
\end{multline}
where it is worth noting that all of the expressions can be easily built from $\mbs{\eta}$.  We will not pursue the symmetry-aware version of the natural parameters, but it would be straightforward to do so using the same approach as the other parameterizations.


\section{Natural Gradient Descent}

In this section we will use our \acp{FIM} to design efficient methods of optimizing functionals of Gaussians.  The idea is to exploit the information geometry of whatever our parameterization is to create a better descent direction than regular gradient descent \citep{amari98}.  

\subsection{\ac{NGD} Derivation}

We begin by assuming that we have a loss functional, $V(q)$, of a Gaussian, $q$, that we wish to minimize.  We assume it can be expressed as the \ac{KL} divergence between our approximation and true Bayesian posterior, $p(\mbf{x}|\mbf{z})$, 
\begin{equation}
V(q) = - \int q(\mbf{x}) \ln \left( \frac{p(\mbf{x|\mbf{z}})}{q(\mbf{x})} \right) \, d\mbf{x},
\end{equation}
where $\mbf{x}$ are the variables to be estimated and $\mbf{z}$ is some data.  If $p(\mbf{x} | \mbf{z})$ is actually Gaussian, we can make $V(q)$ zero by setting $q(\mbf{x}) = p(\mbf{x} | \mbf{z})$.  Usually, $p(\mbf{x} | \mbf{z})$ is not Gaussian due to nonlinearities, but presumably it will be somewhat like a Gaussian in order to approximate it as one.

We will employ a Newton-like optimization scheme to minimize the loss.  Suppose that we start with a Gaussian estimate, $q$, and then make a small change in its parameters to result in $q^\prime$.  The change in the loss functional will be
\begin{equation}
V(q^\prime) - V(q) \approx dV + \frac{1}{2} d^2 V,
\end{equation} 
correct to second order in the differentials.  In terms of parameters, $\mbs{\theta}$, this will be
\begin{equation}
\label{eq:losschange}
V(q^\prime) - V(q) \approx \left( \frac{\partial V}{\partial \mbs{\theta}^T}  \right)^T \delta\mbs{\theta} + \frac{1}{2} \delta\mbs{\theta}^T \underbrace{\left( \frac{\partial^2 V}{\partial \mbs{\theta}^T \partial \mbs{\theta} } \right)}_{\mbox{Hessian}} \delta\mbs{\theta},
\end{equation}
where the Hessian can be expensive to compute.  We can think of this as a quadratic approximation of our loss functional centred at the current estimate, which is minimized by solving
\begin{equation}
\left( \frac{\partial^2 V}{\partial \mbs{\theta}^T \partial \mbs{\theta} } \right) \, \delta\mbs{\theta} = - \left(\frac{\partial V}{\partial \mbs{\theta}^T}\right),
\end{equation}
for $\delta\mbs{\theta}$ then updating $\mbs{\theta}^\prime = \mbs{\theta} + \delta \mbs{\theta}$ and iterating to convergence.  The key step in \acf{NGD} is to approximate the Hessian using the \ac{FIM} in the update scheme so that
\begin{equation}
\label{eq:ngd}
\delta\mbs{\theta} = -\mbs{\mathcal{I}}_{\mbs{\theta}}^{-1} \frac{\partial V}{\partial \mbs{\theta}^T},
\end{equation}
where we can precompute the \ac{FIM} inverse if we like.  We can also simply interpret this as a modification to gradient descent that achieves something more like second-order convergence without any second-order derivatives.  We saw earlier that the \ac{FIM} is exactly the Hessian of the \ac{KL} divergence between two Gaussians.  If $p(\mbf{x} | \mbf{z})$ is `close' to a Gaussian then $V(q)$ will be well approximated by the \ac{KL} divergence between two Gaussians.

Inserting the update back into~\eqref{eq:losschange} we have
\begin{equation}
V(q^\prime) - V(q) \approx - \left( \frac{\partial V}{\partial \mbs{\theta}^T}  \right)^T  \left( \mbs{\mathcal{I}}_{\mbs{\theta}}^{-1} - \frac{1}{2} \mbs{\mathcal{I}}_{\mbs{\theta}}^{-1} \left( \frac{\partial^2 V}{\partial \mbs{\theta}^T \partial \mbs{\theta} } \right)  \mbs{\mathcal{I}}_{\mbs{\theta}}^{-1} \right) \left( \frac{\partial V}{\partial \mbs{\theta}^T}  \right) \approx - \frac{1}{2} \left( \frac{\partial V}{\partial \mbs{\theta}^T}  \right)^T \mbs{\mathcal{I}}_{\mbs{\theta}}^{-1} \left( \frac{\partial V}{\partial \mbs{\theta}^T}  \right) \leq 0,
\end{equation}
which shows that (under the approximations we have made) the loss will decrease (except at a minimum where $\frac{\partial V}{\partial \mbs{\theta}^T} = \mbf{0}$) since the \ac{FIM} is positive definite.

\subsection{Canonical Parameterization - Not Accounting for Symmetry}

We can deconstruct~\eqref{eq:ngd} into the updates for $\mbs{\mu}$ and $\mbs{\Sigma}$:
\begin{equation}
\bbm \delta\mbs{\mu} \\ \vec(\delta\mbs{\Sigma}) \ebm = - \bbm \mbs{\Sigma} & \mbf{0} \\ \mbf{0} & 2 \left(\mbs{\Sigma} \otimes \mbs{\Sigma} \right) \ebm \bbm \frac{\partial V}{\partial \mbs{\mu}^T} \\ \vec\left(\frac{\partial V}{\partial \mbs{\Sigma}}\right) \ebm.
\end{equation}
For the mean we simply have
\begin{equation}
\delta\mbs{\mu} = - \mbs{\Sigma} \frac{\partial V}{\partial \mbs{\mu}^T},
\end{equation}
or rearranging slightly
\begin{equation}
\label{eq:canmean}
\mbs{\Sigma}^{-1} \, \delta\mbs{\mu} = -\frac{\partial V}{\partial \mbs{\mu}^T},
\end{equation}
which looks familiar from \ac{MAP} inference.  For the covariance parameter we have
\begin{equation}
\vec(\delta\mbs{\Sigma}) = - 2 \left(\mbs{\Sigma} \otimes \mbs{\Sigma} \right) \vec\left(\frac{\partial V}{\partial \mbs{\Sigma}}\right),
\end{equation}
which can be turned back into a matrix equation by applying $\mat(\cdot)$ to both sides:
\begin{equation}
\label{eq:delSigma}
\delta\mbs{\Sigma} = -2 \mbs{\Sigma} \frac{\partial V}{\partial \mbs{\Sigma}} \mbs{\Sigma}.
\end{equation}
We will next work through the details, accounting for the symmetric nature of the covariance matrix.

\subsection{Canonical Parameterization - Accounting for Symmetry}

The \ac{NGD} update is now
\begin{equation}
\delta\mbs{\gamma} = -\mbs{\mathcal{I}}_{\mbs{\gamma}}^{-1} \frac{\partial V}{\partial \mbs{\gamma}^T},
\end{equation}
which becomes
\begin{equation}
\bbm \delta\mbs{\mu} \\ \vech(\delta\mbs{\Sigma}) \ebm = - \bbm \mbs{\Sigma} & \mbf{0} \\ \mbf{0} & 2 \mbf{D}^+\left(\mbs{\Sigma} \otimes \mbs{\Sigma} \right)\mbf{D}^{+^T} \ebm \bbm \frac{\partial V}{\partial \mbs{\mu}^T} \\ \vech\left(\frac{\partial V}{\partial \mbs{\Sigma}}\right) \ebm.
\end{equation}
The mean update is the same as the previous section.  For the covariance update we have
\begin{equation}
\label{eq:delgamma2}
\vech(\delta\mbs{\Sigma}) = -2 \mbf{D}^+ \left(\mbs{\Sigma} \otimes \mbs{\Sigma}\right)\mbf{D}^{+^T} \, \vech\left(\frac{\partial V}{\partial \mbs{\Sigma}}\right).
\end{equation}
We note that
\begin{equation}
\vech\left(\frac{\partial V}{\partial \mbs{\Sigma}}\right) = \mbf{D}^T \vec\left( \frac{\partial V}{\partial \mbs{\Sigma} }\right),
\end{equation}
and 
\begin{equation}
\vech(\delta\mbs{\Sigma}) =  \mbf{D}^+ \vec(\delta\mbs{\Sigma}).
\end{equation}
Inserting these last two equations into~\eqref{eq:delgamma2} we have
\begin{equation}
\mbf{D}^+ \vec(\delta\mbs{\Sigma}) = -2 \mbf{D}^+ \left(\mbs{\Sigma} \otimes \mbs{\Sigma}\right)\underbrace{\mbf{D}^{+^T} \mbf{D}^T}_{\mbf{D} \mbf{D}^+ \;\mbox{\scriptsize by \eqref{eq:dup2}}} \vec\left( \frac{\partial V}{\partial \mbs{\Sigma}} \right).
\end{equation}
Premultiplying both sides by $\mbf{D}$ we have
\begin{equation}
\underbrace{\mbf{D}\mbf{D}^+ \vec(\delta\mbs{\Sigma})}_{\vec(\delta\mbs{\Sigma}) \;\mbox{\scriptsize by~\eqref{eq:dup3}}} = -2 \underbrace{\mbf{D}\mbf{D}^+ \left(\mbs{\Sigma} \otimes \mbs{\Sigma}\right)\mbf{D}}_{\left(\mbs{\Sigma} \otimes \mbs{\Sigma}\right)\mbf{D} \;\mbox{\scriptsize by~\eqref{eq:dup4}}} \mbf{D}^+ \vec\left( \frac{\partial V}{\partial \mbs{\Sigma}} \right),
\end{equation}
then
\begin{equation}
\vec(\delta\mbs{\Sigma}) = -2 \left(\mbs{\Sigma} \otimes \mbs{\Sigma}\right) \underbrace{\mbf{D}\mbf{D}^+ \vec\left( \frac{\partial V}{\partial \mbs{\Sigma}} \right)}_{\vec\left( \frac{\partial V}{\partial \mbs{\Sigma}} \right) \;\mbox{\scriptsize by~\eqref{eq:dup3}}},
\end{equation}
and then
\begin{equation}
\vec(\delta\mbs{\Sigma}) = -2 \left(\mbs{\Sigma} \otimes \mbs{\Sigma}\right) \vec\left( \frac{\partial V}{\partial \mbs{\Sigma}} \right).
\end{equation}
Applying the $\mat(\cdot)$ operator to both sides we finally have
\begin{equation}
\label{eq:cancov}
\delta\mbs{\Sigma} = -2 \mbs{\Sigma}  \frac{\partial V}{\partial \mbs{\Sigma}} \mbs{\Sigma},
\end{equation}
which amazingly is exactly the update in~\eqref{eq:delSigma} that we arrived at when we did not account for the symmetric nature of $\mbs{\Sigma}$.

\subsection{Hybrid Parameterization - Not Accounting for Symmetry}

The \ac{NGD} update for our symmetry-blind hybrid parameters, $\mbs{\alpha}$ is
\begin{equation}
\delta \mbs{\alpha} = - \mbs{\mathcal{I}}_{\mbs{\alpha}}^{-1}  \frac{\partial V}{\partial \mbs{\alpha}^T}.
\end{equation}
Working through the same logic as the symmetry-blind canonical section we have
\begin{subequations}
\begin{eqnarray}
\mbs{\Sigma}^{-1} \, \delta\mbs{\mu} & = & -\frac{\partial V}{\partial \mbs{\mu}^T}, \\
\delta\mbs{\Sigma}^{-1} & = & -2 \mbs{\Sigma}^{-1}  \frac{\partial V}{\partial \mbs{\Sigma}^{-1}} \mbs{\Sigma}^{-1}. \label{eq:delSigmaInv}
\end{eqnarray}
\end{subequations}
The first equation is the same as~\eqref{eq:canmean} while the second equation is not equivalent to~\eqref{eq:cancov}.

\subsection{Hybrid Parameterization - Accounting for Symmetry}

The \ac{NGD} update for our symmetry-aware inverse covariance parameters, $\mbs{\beta}$ is
\begin{equation}
\delta \mbs{\beta} = - \mbs{\mathcal{I}}_{\mbs{\beta}}^{-1}  \frac{\partial V}{\partial \mbs{\beta}^T}.
\end{equation}
Working through the same logic as the symmetry-aware canonical section we have
\begin{subequations}
\begin{eqnarray}
\mbs{\Sigma}^{-1} \, \delta\mbs{\mu} & = & -\frac{\partial V}{\partial \mbs{\mu}^T}, \\
\delta\mbs{\Sigma}^{-1} & = & -2 \mbs{\Sigma}^{-1}  \frac{\partial V}{\partial \mbs{\Sigma}^{-1}} \mbs{\Sigma}^{-1}.
\end{eqnarray}
\end{subequations}
Again, these are the same updates we arrived at not accounting for symmetry in the previous section.

\subsection{Choosing the Hybrid Parameterization}

Noticing that for our particular choice of $V$ that \citep{opper09, barfoot_ijrr20}
\begin{subequations}
\begin{eqnarray}
\frac{\partial V}{\partial \mbs{\mu}^T} & = &  \mbs{\Sigma}^{-1} \mathbb{E}_q[ (\mbf{x} - \mbs{\mu}) \phi(\mbf{x})],  \label{eq:deriv1a}   \\
\frac{\partial^2 V}{\partial \mbs{\mu}^T \partial \mbs{\mu}} & = & \mbs{\Sigma}^{-1} \mathbb{E}_q[ (\mbf{x} - \mbs{\mu}) (\mbf{x} - \mbs{\mu})^T \phi(\mbf{x})] \mbs{\Sigma}^{-1}  - \mbs{\Sigma}^{-1} \,\mathbb{E}_q[\phi(\mbf{x})], \label{eq:deriv1b}  \\
\frac{\partial V}{\partial \mbs{\Sigma}^{-1}} & = &  -\frac{1}{2}\mathbb{E}_q[ (\mbf{x} - \mbs{\mu}) (\mbf{x} - \mbs{\mu})^T\phi(\mbf{x})]   + \frac{1}{2}  \mbs{\Sigma} \,\mathbb{E}_q[\phi(\mbf{x})] + \frac{1}{2} \mbs{\Sigma}, \label{eq:deriv2}
\end{eqnarray}
\end{subequations}
where $\phi(\mbf{x}) = -\ln p(\mbf{x}, \mbf{z})$, we have the nice relationship
\begin{equation}
\label{eq:derivRelation}
\frac{\partial V}{\partial \mbs{\Sigma}^{-1}} = \frac{1}{2} \mbs{\Sigma} - \frac{1}{2} \mbs{\Sigma} \,\frac{\partial^2 V}{\partial \mbs{\mu}^T \partial \mbs{\mu}} \,\mbs{\Sigma}.
\end{equation}
Plugging~\eqref{eq:derivRelation} into our canonical covariance update from~\eqref{eq:delSigma} we have
\begin{equation}
\delta\mbs{\Sigma} = -\mbs{\Sigma}^3 + \mbs{\Sigma}^3 \, \frac{\partial^2 V}{\partial \mbs{\mu}^T \partial \mbs{\mu}} \, \mbs{\Sigma}^3,
\end{equation}
which does not appear to be an improvement.  

However, plugging~\eqref{eq:derivRelation} into our hybrid covariance update in~\eqref{eq:delSigmaInv} we have
\begin{equation}
\delta\mbs{\Sigma}^{-1} = - \mbs{\Sigma}^{-1} + \frac{\partial^2 V}{\partial \mbs{\mu}^T \partial \mbs{\mu}},
\end{equation}
which conveniently removes the extra $\mbs{\Sigma}^{-1}$ matrices.  Since $\delta\mbs{\Sigma}^{-1}$ is the change in the inverse covariance, we can express this update simply as
\begin{equation}
\mbs{\Sigma}^{-1}  \leftarrow  \frac{\partial^2 V}{\partial \mbs{\mu}^T \partial \mbs{\mu}},
\end{equation}
which is quite convenient to implement.

We can then express the complete hybrid update neatly as
\begin{subequations}
\label{eq:hybridupdate}
\begin{eqnarray}
\mbs{\Sigma}^{-1} \, \delta\mbs{\mu} & = & - \frac{\partial V}{\partial \mbs{\mu}^T}, \\
\mbs{\Sigma}^{-1} & \leftarrow & \frac{\partial^2 V}{\partial \mbs{\mu}^T \partial \mbs{\mu}}, \\
\mbs{\mu} & \leftarrow & \mbs{\mu} + \delta\mbs{\mu}, 
\end{eqnarray}
\end{subequations}
where the derivatives of $V$ will require the current estimate of $q$ to be evaluated.  However, if we cycle through them from top to bottom, we will naturally update both $\mbs{\mu}$ and $\mbs{\Sigma}^{-1}$.  

\subsection{Hybrid Update Preserves Sparsity}

The hybrid \ac{NGD} update also has the nice property of preserving the sparsity inherent in the inverse covariance matrix \citep{barfoot_ijrr20}. If we assume that the joint likelihood, $p(\mbf{x},\mbf{z})$, factors then we can express the loss functional as
\begin{eqnarray}
V(q) & = & \mathbb{E}_q[ -\ln p(\mbf{x}|\mbf{z})] + \mathbb{E}_q[ \ln q(\mbf{x})] \nonumber \\
& = & \mathbb{E}_q[-\ln  p(\mbf{x},\mbf{z})] + \underbrace{\mathbb{E}_q[ \ln p(\mbf{z})]}_{\mbox{constant}} + \frac{1}{2} \ln \left( | \mbs{\Sigma}^{-1} | \right) \nonumber \\
& = & \mathbb{E}_q\left[-\ln  \left( \prod_{k=1}^K p(\mbf{x}_k,\mbf{z}_k) \right)\right] + \frac{1}{2} \ln \left( | \mbs{\Sigma}^{-1} | \right) + \mbox{constant} \nonumber \\
& = & \sum_{k=1}^K \underbrace{\mathbb{E}_{q_k}\left[-\ln p(\mbf{x}_k,\mbf{z}_k) \right]}_{V_k(q_k)} + \underbrace{\frac{1}{2} \ln \left( | \mbs{\Sigma}^{-1} | \right)}_{V_0} + \; \mbox{constant} \nonumber \\
& = & V_0 + \sum_{k=1}^K V_k(q_k) + \mbox{constant},
\end{eqnarray} 

where
\begin{equation}
\mbf{x}_k = \mbf{P}_k \mbf{x}
\end{equation}
is a subset of the variables associated with the $k$th factor and $\mbf{P}_k$ an appropriate projection matrix.  The first derivative of the loss is
\begin{equation}
\frac{\partial V}{\partial \mbs{\mu}^T} = \underbrace{\frac{\partial V_0}{\partial \mbs{\mu}^T}}_{\mbf{0}} + \frac{\partial}{\partial \mbs{\mu}^T} \sum_{k=1}^K V_k = \sum_{k=1}^K \frac{\partial V_k}{\partial \mbs{\mu}^T} = \sum_{k=1}^K \mbf{P}_k^T \frac{\partial V_k}{\partial \mbs{\mu}_k^T},
\end{equation}
where in the last step, the derivative simplifies to being over just $\mbs{\mu}_k$ since that term only depends on $q_k$.  We have a similar thing for the second derivative:
\begin{equation}
\mbs{\Sigma}^{-1} = \frac{\partial^2 V}{\partial \mbs{\mu}^T \partial \mbs{\mu}} = \underbrace{\frac{\partial^2 V_0}{\partial \mbs{\mu}^T\partial \mbs{\mu}}}_{\mbf{0}} + \frac{\partial^2}{\partial \mbs{\mu}^T\partial \mbs{\mu}} \sum_{k=1}^K V_k = \sum_{k=1}^K \frac{\partial^2 V_k}{\partial \mbs{\mu}^T\partial \mbs{\mu}} = \sum_{k=1}^K \mbf{P}_k^T \frac{\partial^2 V_k}{\partial \mbs{\mu}_k^T\partial \mbs{\mu}_k} \mbf{P}_k.
\end{equation}
From the right-most expression, we see that the inverse covariance will always maintain the same sparsity pattern.  From here, \citet{barfoot_ijrr20} show how to use this update scheme to carry out large-scale inference for problems in robotics.

\section{Conclusion}

We have derived the \ac{FIM} and \ac{NGD} updates for a handful of different multivariate Gaussian representations.  The take away message is that the hybrid parameterization (comprising mean and inverse covariance) seems to offer advantages over the others and our update in~\eqref{eq:hybridupdate} inherently takes care of the symmetric (and sparse) nature of $\mbs{\Sigma}^{-1}$.  Please refer to \citet{barfoot_ijrr20} for an application of this approach to multivariate Gaussian variational inference.

\bibliographystyle{asrl}
\bibliography{refs}

\end{document}